\documentclass[letterpaper, 10 pt, conference]{ieeeconf}  

\IEEEoverridecommandlockouts                              

\usepackage{graphicx} 
\graphicspath{{data/}{images/}}
\usepackage{subfig}
\usepackage{amsmath} 
\usepackage{amssymb}  
\usepackage{multirow}
\usepackage{url}

\title{\LARGE \bf
Assessing virtualization effects in simulations of distributed robotics
}

\author{Sekou L. Remy
\thanks{Sekou L. Remy is with IBM Research Africa,
        Nairobi, Kenya
        {\tt\small sekou@ke.ibm.com}}%
}

\begin{document}

\maketitle
\thispagestyle{empty}
\pagestyle{empty}

\begin{abstract}
In this work, our aim is to identify whether the choice of virtualization strategy influences the performance of simulations in robotics.
Performance is quantified in the error between a reference trajectory and the actual trajectory for the ball moving along the surface of a smooth plate.
The two-sample Kolmogorov-Smirnov test is used to assess significance of variations in performance under the different experimental settings.
Our results show that the selection of virtualization technology does have a significant effect on simulation, and moreover this effect can be amplified by the use of some operating systems.
While these results are a strong cause for caution, they also provide reason for optimism for those considering ``repeatable robotics research'' using virtualization.

\end{abstract}

\section{INTRODUCTION}

Realistic computer based simulation has been an effective tool -- in both academia \cite{618020,corke2011robotics} and industry \cite{ma1997mdsf,lerner2010does} -- to explore content that is difficult to recreate in the physical world (safety, distance), challenging to understand because of scale, or even impossible to occur under easily reproducible conditions.
Simulation environments themselves can sometimes also be difficult to install, configure, and deploy, and if there is variability in any of these phases then it is possible for the results that are obtained to be influenced in a unexpected way.
Simulation technology can benefit from advances in computing and information technology, but only if these advances can be appropriately incorporated into the workflow.

In this work, we evaluate the impact of widely used abstractions of computing hardware and subsystems when running simulations of physically based robotics simulation environments.
This type of simulation is important as it includes distributed robots and also emulates the networks that facilitate their control.
In this simulation setting, onboard sensors are collecting data on simulated robots which then transmit this data over an emulated network to another node which calculates the desired control values that the simulated actuators execute.
This entire simulation system is deployed leveraging \textit{virtualization}.
Containers and Virtual Machines (VMs) are the two types of virtualization considered in this work.
They are widely used to deploy software, and we present results which begin to characterize the impact of their use in robotics.
This work is critical because of the number of users of the technologies involved.
With over 500K pulls of ROS Containers and use of VMs since ROS Fuerte in 2012, whether in the cloud or on the desktop, virtualization's penetration is deep and the impact on experimental results must be fully understood.

We wish to test the null hypothesis $H_0$:
the observed performance of a controller is consistent in all deployment environments that a physics-based simulation is performed.
Performance is quantified by performing a distributed control task using an emulated network.
This investigation is important as the research community is embracing virtualization, so we need to ensure that the validity of experimental findings generated is not compromised by inconsistencies that may arise from the use of virtualization approaches.
The main technical contributions of this paper are the implementation of the robotic simulation environment in a new network emulation framework, the deployment of the simulation, the experimental assessment of the performance in these simulation, and the statistical evaluation of this performance.

\section{BACKGROUND}
Hardware virtualization is a software abstraction that creates a working representation of a computer system's hardware \cite{goldberg1974survey, creasy1981origin}.
Initially developed to correct shortcomings in computer architectures of the time, one use case for the technology was to provide extended ``machines'' so that user programs could be executed in the presence of dual state hardware which had a \textit{priviledged} and \textit{non-priviledged} mode.
Another use case was to permit programs written for special purpose hardware to be compiled and executed on general purpose ones instead.

While a great deal has changed since the 1960's when it was first developed, virtualization is still very present in modern computing.
In fact, virtualization is in the midst of a significant resurgence fueled by the architectural extensions which major microprocessor manufacturers have provided\cite{adams2006comparison}, as well as its integral role in Cloud Computing.
Virtual machines (VMs) and containers are two competing virtualization approaches which have been developed.
Virtual machines differ from containerized virtualization in that each application which runs in a VM is executed on its own operating system (i.e. it has its own kernel) and executes on its own virtualized copy of hardware which is provided by a hypervisor.
This hypervisor software thus provides isolation for virtual machines running on physical hosts \cite{7164727}.
In contrast, containerization does not employ the use of a hypervisor, and instead uses a container engine which translates the instructions from the guest machine to the host machine's kernel.
If there are multiple containers operating on the same host, they share the same kernel.
Isolation between processes is provided through the use of namespaces that the Linux operating system provides \cite{biederman2006multiple}.
These namespaces enable \textit{lightweight} virtualization, as devices can now effectively be shared without collision of the processes and applications performing work at the same time.

Virtualization makes computing resources easier to access, and easier to share; And these are two of the challenges that those doing research and development in robotics experience.
As an example, in the words of the Open Source Robotics Foundation, curators of the Gazebo simulation environment:
\begin{quote}
With the advancements and standardization of software containers, roboticists are primed to acquire a host of improved developer tooling for building and shipping software.
To help alleviate the growing pains and technical challenges of adopting new practices, we have focused on providing an official resource for using Gazebo with these new technologies.\footnote{https://hub.docker.com/\_/gazebo/}
\end{quote}

The previous quote applies to containerization, however since 2012 Gazebo has also been included on preconfigured VMs, and released with ROS distributions.
Moreover, over the past two years, key robotics conferences like ICRA have held workshops and tutorials teaching researchers how to use virtualization\footnote{http://the.swarming.buzz/ICRA2017/virtual-machine-tutorial/}, and papers are frequently shared with deployable environments for others to evaluate\footnote{http://d3s.mff.cuni.cz/software/ros\_sCPS\_testbed/}.
Virtualization is clearly addressing a need, and as it is being increasingly deployed its influence on the work done in the field is growing.

The goal of this work is to identify whether the choice of virtualization strategy influences the performance of simulations in robotics, thus influencing the experiments that are done with virtualization.
We are specifically focus on simulations which incorporate realistic networking settings.
The ability to identify the effect of virtualization (if any), would be a precursor to developing models which could map research findings from one experimental setting and transform them to findings from another.
The development of such models would enable simulations deployed in virtualized environments to provide more useful virtual testbeds for science, instead of being relegated to system integration frameworks or for use as one off deployment platforms.

Many existing multi-robot simulators like Gazebo, Webots, and MORSE do not provide advanced communication models \cite{kudelski2013robonetsim}, so researchers have developed extensions to integrate network simulations and emulations with their robot simulations \cite{ye2001evaluating, matena2016model}.
In our approach
we choose \textbf{mininet} based emulation because it readily permits existing robot control codebases to be used without modification.
Extensions to mininet also permit mobility.
Mininet is a system for rapidly prototyping large networks on a single computer which supports running unmodified code on emulated machines \cite{lantz2010network}.
Mininet uses containers and allows processes to be placed in namespaces with their own properties like IP addresses.
Developed in the Software Defined Networking community, mininet makes use of OpenVSwitch for switching and uses NetEm, a part core Linux subsystem, to emulate the links between virtual hosts (aka the processes running in namespaces).

Conceptually, there are 5 distinct layers that are essential to the evaluations in this work.
The lowest layer, Layer 1 is the hardware.
This layer provides the resource that the operating system (Layer 2) will run upon.
It is possible for this hardware to be provided through virtualization since we are focused on the effect of virtualization on simulation which is in a higher layer.
Layer 2 provides the host operating system that the virtualization environment will be executed.
The operating system needs to be able to support virtualization either by permitting a program to run, or by providing access to its subsystem functionality.
Layer 3 captures the two types of virtualization, VMs and containers.
Layers 4 and 5 contain mininet and the network simulation codebase that we will use to characterize the effect of virtualization.
Since Layer 3 could be influenced by the layers that enable it, any experiments will have to consider the cumulative effects of the combined substrate.

\begin{table}[t]
\caption{Virtualization centric layers for robot simulations}
\label{5layers}
\begin{center}
\begin{tabular}{|c|c|}
\hline
L5 & Robot simulations\\
\hline
L4 & Mininet\\
\hline
L3 & Virtualization\\
\hline
L2 & Host OS\\
\hline
L1 & Host machine\\
\hline
\end{tabular}
\end{center}
\end{table}

\section{EXPERIMENTAL SETUP}
The experimental infrastructure for this work is captured in the configurations presented in Table \ref{exps}.
In this table nine combinations of base \text{hardware}, operating systems, and virtualization options.
These configurations of infrastructure represent those that can readily be found in the robotics community.

\begin{table}[t]
\caption{Scenarios a-i are the 9 possible experimental configurations considered in this work.
These scenarios provide different combinations of hardware, operating systems, and virtualization for comparison of the performance of the robotic simulation.}
\label{exps}
\begin{center}
\begin{tabular}{|c|c|c|c|c|}
\hline

L1 & L2 & \multicolumn{3}{c|}{L3}\\\cline{3-5}
& & Docker& Native& VirtualBox\\
\hline\hline
i686 & Ubuntu & a & b & c\\ 
Intel i7 & OSX & d & e & f\\
ARM64 QEMU& Ubuntu & g & h & i\\

\hline
\end{tabular}
\end{center}
\end{table}

\subsection{Distributed Robotic Control}
\subsubsection{Robot}
The robot, or \textit{plant}, selected for this work, the \textit{ball on plate}, is widely studied in control theory and it is an example of a dynamic system with an unstable equilibrium point.
In this section we describe the system, first by introducing the \textit{ball on beam} system, and then we continue to introduce the feedback controller we use.

For this plant, the position $x$ of the ball is controlled by changing the angle $Y$ of the beam which causes the ball to roll along its surface.
By controlling the orientation of the beam, the position of the ball is thus indirectly changed.
To maintain any target ball position, the controller must actively operate since the ball's position is affected by gravity at all times.
The ball and plate system is modeled as an extension of the ball on beam to two dimensions.
This extension is rooted in the assumption that the motion in each dimension of the plane defined by the plate can be controlled independently.
That assumption is valid for small Euler angles under the linearized dynamics.

Consequently, controlling the position of the ball involves determining the input for two actuators that adjust the Euler angles of the plate's rotational axes.
In this work the state of the system is partially observable, and only the location of the ball on the plate, and the roll and pitch angles of the plate are known.
The reader is referred to \cite{awtar2002mechatronic} for the complete derivation of the equations of motion used to model the system.
In this work, the ball on plate system is modeled as two independent ball on beam systems, so equations presented for only one dimension of the system.
\begin{equation}
\frac{A(s)}{U(s)} = \frac{1}{s(b_1 s+b_0)} 
\label{2.2.11}
\end{equation}
Using (\ref{2.2.11}) the relationship between the input voltage to the motor ($U$) and the angle of the motor shaft ($A$) can be calculated.
In this work, the published values of $b_1 = 0.01176$ and $b_0=0.58823$ \cite{randolph1998design} are applied.
These values account for the physical properties of the system.
They incorporate the total load inertia, the total load friction, the gear ratio for the motor, and assume negligible armature inductance.

\begin{figure}[t]
\centering{
\includegraphics[width=.9\columnwidth]{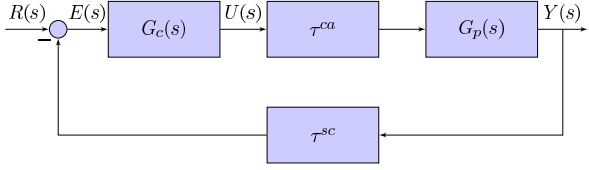}
}
\caption{Control of the ball on plate example.
The input to the system, $R(s)$, defines the target ball trajectory.
$E(s)$ captures the error from that target and this data is used as input to the controller.
The controller then generates the control input to the plant, $U(s)$ and uses the switch to transmit this data to the robot (See Fig. \ref{topologyfig}).
The output of the plant, $Y(s)$ is transmitted back through the switch to the controller.
In this figure, $\tau^{ca}$ is the delay between controller and actuator, and $\tau^{sc}$ is the delay between sensor and controller.
}
\label{fig:controlloop}
\end{figure}

A linear mapping (\ref{linearmapp}) is assumed between the Euler angle of the beam ($Y$) and the angle of the motor shaft ($A$).
This linear relationship is only valid for small angles (values between $-32$ and $32$ degrees), but it results in the simplified equation for (\ref{2.2.2_new}), which is the transfer function for $G_{P1}(s)$ - the effect of control input on the beam orientation.
\begin{eqnarray}
\frac{Y(s)}{A(s)} &=& \frac{1}{16} \label{linearmapp} \\
G_{P1}(s) &=& \frac{Y(s)}{U(s)} = \frac{1}{16 s (b_1 s+b_0)} 
\label{2.2.2_new}
\end{eqnarray}
The transfer function between the Euler angle and the ball position is given by (\ref{2.2.6}).
This transfer function was obtained by linearizing the equations of motion for the system about an Euler angle and an angular rate of zero.
This model assumes that the ball rolls without slip, and with negligible friction along the surface.
\begin{equation}
G_{P2}(s) = \frac{X(s)}{Y(s)} = -\frac{7}{s^2}
\label{2.2.6}
\end{equation}
Ideally, to better simulate the plant, $G_{P1}(s)$ should update at a much faster rate than $G_{P2}(s)$, however in this work they are updated at the same rate and at the same time step.
As such, the effective system plant, $G_{P}(s)$=$G_{P1}(s)*G_{P2}(s)$.
\subsubsection{Robot controllers}
For each axis of the system, the position of the ball, and the Euler angle of the plate were each controlled with one PD controller.
This results in four controllers of the form presented in (\ref{eq:pd}).
\begin{equation}
G_{C}(s) = \frac{K_d s^2 + K_p s }{s}
\label{eq:pd}
\end{equation}

\subsubsection{Networking}
Communication between the  \textit{controller} and the \textit{plant} was implemented using the Robot Operating System (ROS) \cite{quigley2009ros}.
ROS provides a structured communications layer above the host operating systems, and is an open-source project widely used in the robotics community \cite{curran2015evaluating}
The controller and the plant were run on hosts $sta1$ and $sta2$ respectively, while the rosmaster node was run on a third host, $h1$.
The ROS publish and subscribe paradigm is used to connect these nodes, and leveraged TCP to provide message delivery.

\begin{figure}[!t]
\centering
\includegraphics[width=\columnwidth]{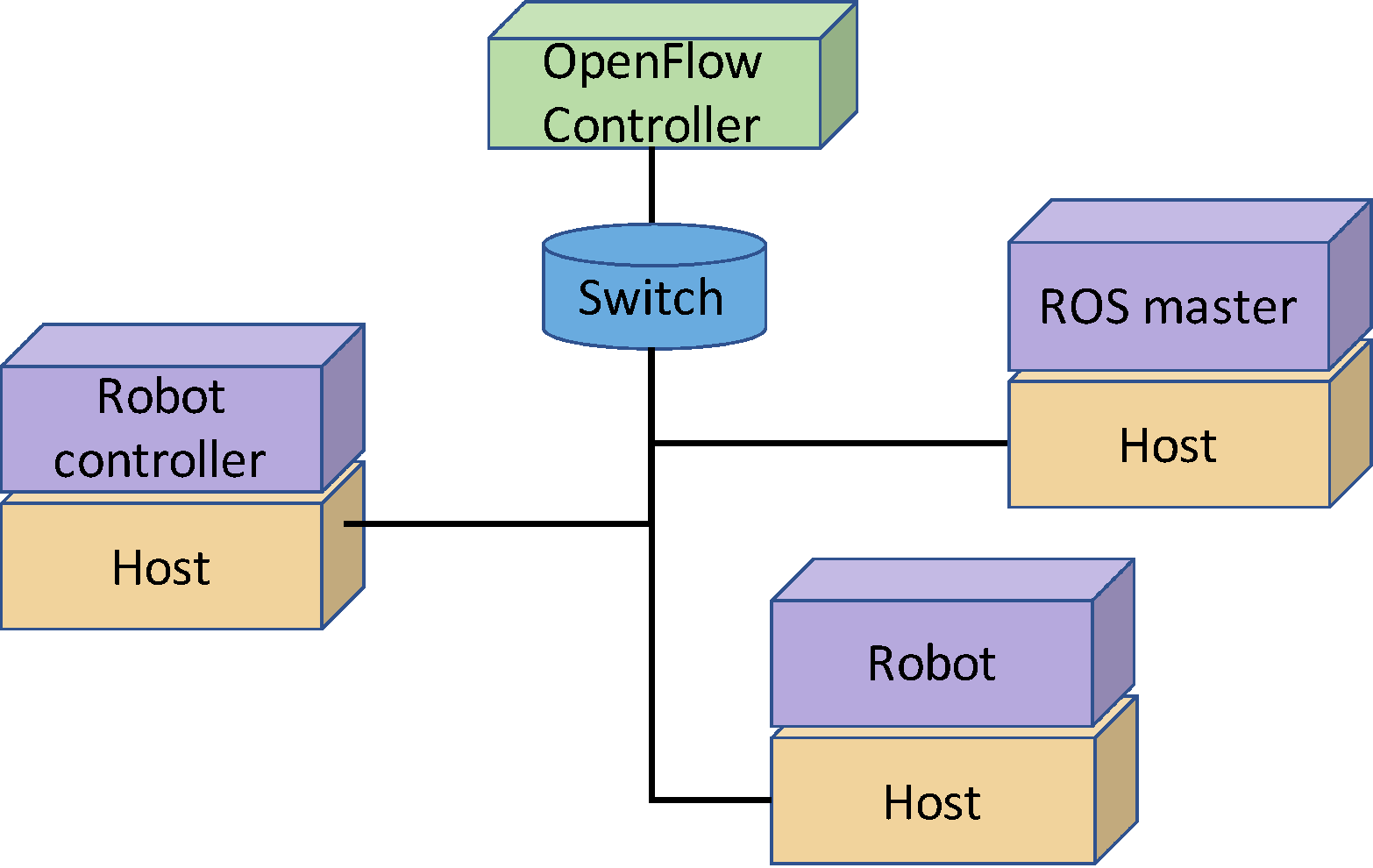}
\caption{
An OpenFlow controller manages the transfer of IP packets flowing through the network switch.
This topology forms layers 4 \& 5 of the model presented in Tab. \ref{5layers}.
This is the communication substrate that permits the robot controller to  publish data to, and subscribe to data from the robot.
}
\label{topologyfig}
\end{figure}

The three hosts and the network that connected them were emulated using mininetAPI.
As shown in Fig. \ref{topologyfig}, these hosts were connected to the same switch, and the attributes for each link were: delay = $10$ms;  jitter = $1$ms; loss rate = $1$\%; bandwidth = $10$Mb.
The reader should note that this results in a $40$ms round-trip communication time between each pair of hosts.

\subsubsection{Instrumentation}
To characterize performance of the control task, a log is generated by the controller process which captures the control action and the subsequent plant state.
Here the state is defined as the position of the ball and the orientation of the plate.
Each of these properties (state and control) are timestamped by the controller application.
Concurrently, the \textit{tcpdump} utility is used to log all network traffic on each of the virtual hosts.
This utility permits introspection of the lower level properties of the network traffic.
Since all the virtual hosts share the same system clock of the underlying OS, their TCP traces are synchronized and there is complete visibility of the networking traffic.

\subsection{Experiments}
The experiments were run in three settings: a Toshiba Satellite computer, an Apple Macbook Pro, and on a cloud-hosted virtual machine (QEMU).
To collect data, vitualization from either a docker container or a VirtualBox VM was installed, and code from a repository was checked out.
This code contained python source that configured the mininetAPI and launches the three ROS nodes.
There were two reference trajectories used in the control task, one for the X- and Y- axis respectively.
Both trajectories were periodic, with $T = 100s$.
The error between the reference and the actual trajectory for the ball along each axis is the core performance metric for this work.
This `virtual testbed' was created to run for four contiguous $6000$ seconds windows creating data in separate folders.

\begin{figure}[!t]
\centering
\includegraphics[width=.8\columnwidth]{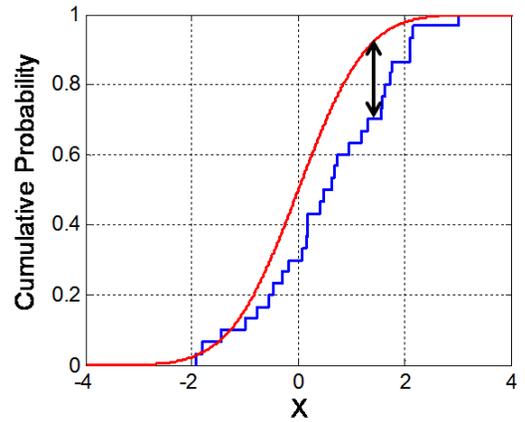}
\caption{Graph depicting an example of the Kolmogorov-Smirnov test.}
\label{kstest}
\end{figure}

So the aim will be to uncover whether there is a statistically significant difference in the errors observed when performing control under the various conditions.
Significance will be evaluated using the two-sample Kolmogorov-Smirnov (KS) test \cite{metchev2002two}.
In this test, the largest difference between two considered sample distributions are assessed to determine if the two samples could have been drawn from the same distributions (See Fig. \ref{kstest}).
The smaller the difference, the higher the likelihood that the samples are from the same distribution.

\noindent Note: Since Docker does not run on the $i686$ chipset, and mininet cannot run on OSX, data cannot be collected for Scenarios -a and -e;
Data will only be collected for the remaining seven scenarios in Tab. \ref{exps}.

\section{RESULTS}
If the null hypoothesis is trues, then there is no statistically significant difference in performance observed in the various experimental settings considered.
This would mean that no matter what virtualization was used to enable simulation, similar results would be observed, and thus truly repeatable robotics simulations would be performed.

\begin{figure}[t]
\centering
\includegraphics[width=\columnwidth]{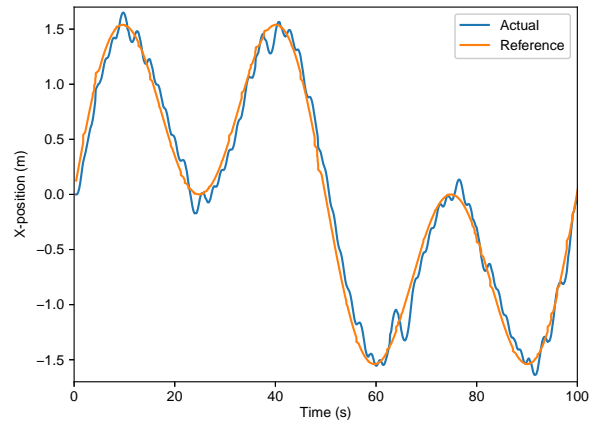}
\caption{One period of the position of the ball when controlled without virtualization (Scenario-b from Tab. \ref{exps}).}
\label{b-position}
\end{figure}

\begin{figure}[t]
\centering
\includegraphics[width=\columnwidth]{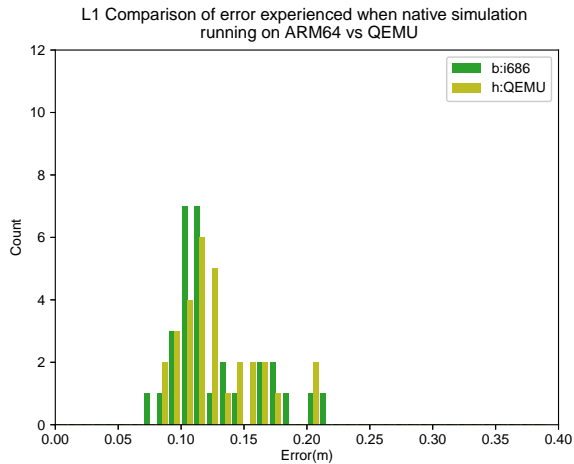}
\caption{Histograms of presenting the performance errors observed in a pair of scenarios considered in this work.
Each of the histograms are asymmetric, and have a non-zero mean}
\label{fig:L1-bh}
\end{figure}

\begin{figure*}
\centering
\subfloat[][Failed to reject]{
\includegraphics[width=.45\textwidth]{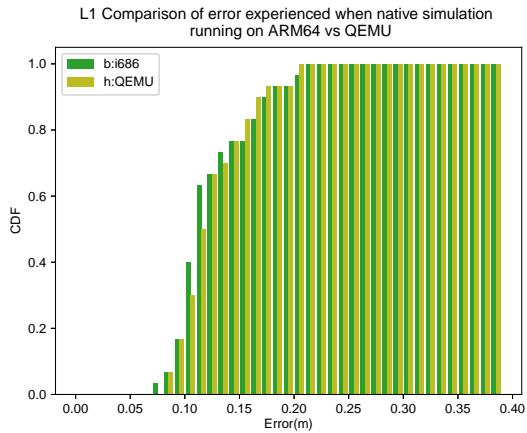}
\label{fig:cdfL1-bh}
}\hfill
\subfloat[][Rejected]{
\includegraphics[width=.45\textwidth]{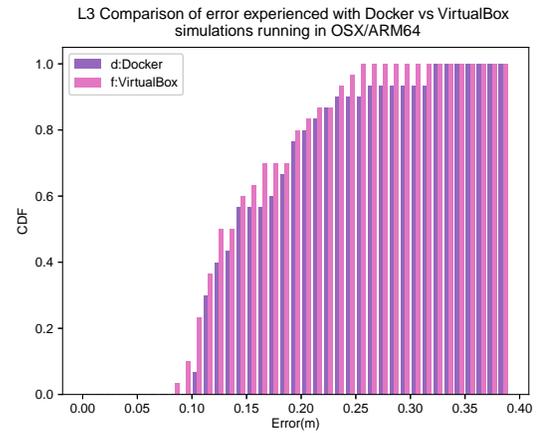}
\label{fig:cdfL1-df}
}
\caption{Cumulative distribution functions for the performance errors observed in two pairs of scenarios considered in this work.
In one comparison (comparing Docker and VirtualBox), the null hypothesis was rejected, i.e. there was enough evidence to conclude that the two samples were not drawn from the same distribution.
(See Tabs. \ref{selfsim}-\ref{comparsions2})
}
\label{fig:cdf}
\end{figure*}

Fig. \ref{b-position} presents the target and actual trajectory for a controller running using Mininet Python API on an Ubuntu host laptop (Scenario-b).
These figures show that control is being performed, however there is error.
This error is averaged over the duration of the period (100$s$) to derive a measure of the performance over the course of the experiement.
In Fig. \ref{fig:L1-bh} data from Scenario-b are presented as a histogram of the error with error data from Scenario-h (with one of the four runs from each scenario).
Visually, the histograms appear similar, however results from the application of the two-sample KS test will assess this formally.
Fig. \ref{fig:cdf} presents two pairs of the cumulative distribution functions highlighting the KS test in action.

\begin{table}[t]
\caption{Pairwise comparison of the 4 samples of each scenario. Comparisons are evaluated via the Kolmogorov-Smirnov test and significance assesed at $p<0.001$. The data show that some scenarios have more consistency than others. When mininet is running with either VirtualBox or Docker there is at least one comparison where the null hypothesis is rejected. $N=6$}
\label{selfsim}
\begin{center}
\begin{tabular}{|c|c|c|c|}
\hline
L1&L2 & L3&Rejected\\
\hline

i686&Ubuntu&- & 0 \\
i686&Ubuntu&VirtualBox & 3 \\
Intel i7&OSX&Docker  & 3 \\
Intel i7&OSX&VirtualBox & 3 \\
QEMU&Ubuntu&Docker & 1 \\
QEMU&Ubuntu&- & 0 \\
QEMU&Ubuntu&VirtualBox & 0 \\

\hline
\end{tabular}
\end{center}
\end{table}

\begin{table}[tp]
\caption{Pairwise comparison of the 4 samples of each scenario.
Comparisons are evaluated via the Kolmogorov-Smirnov test and significance assesed at $p<0.001$.
The data show that significant differences between Docker and VirtualBox do occur, and suggests that they may be dependent on the infrastructure used.
$N=16$.
}
\label{comparsions}
\begin{center}
\begin{tabular}{|c|c|c|c|}
\hline
&Scenario-1 & Scenario-2 & H0 Rejected\\
\hline
Exp. control   &i686/Ubuntu & ARM64/QEMU & 0 \\
\hline
\hline
OSX&Docker & VirtualBox & 9 \\
\hline
Ubuntu& Docker & VirtualBox & 1 \\

\hline
\end{tabular}
\end{center}
\end{table}

\begin{table}[tp]
\caption{Pairwise comparison of the 4 samples of each scenario to 4 samples from the experimental controls.
Comparisons are evaluated via the Kolmogorov-Smirnov test and significance assesed at $p<0.001$.
The data show that there are significant differences that commonly occur based on the run and on the infrastructure used.
Also, full virtualization provides performance more consistent with hardware than lightweight virtualization.
$N=16$.
}
\label{comparsions2a}
\begin{center}
\subfloat[][Control 1: Number of rejections]{
\begin{tabular}{|c|c|c|}
\hline
&OSX & Ubuntu\\
\hline
VirtualBox& 0 & 0 \\
\hline
Docker&10&2\\
\hline
\end{tabular}}

\subfloat[][Control 2: Number of rejections]{
\begin{tabular}{|c|c|c|}
\hline
&OSX & Ubuntu\\
\hline
VirtualBox& 1 & 0 \\
\hline
Docker&7&0\\
\hline
\end{tabular}}
\label{comparsions2}
\end{center}
\end{table}
We first consider the consistency in each of the scenarios.
Since each scenario was run four times, there are 6 pairs of comparisons for each scenario set.
Tab. \ref{selfsim} shows the results of the KS test for each of the scenarios from Tab. \ref{exps}.
This table shows that there were no statistically significant differences in performance of controllers when running natively on Ubuntu on the Toshiba laptop, nor when running on the QEMU VM (Scenarios -b and -h respectively).
Even when VirtualBox was run on the QEMU VM, there was no statistically significant difference in performance observed (Scenario-i).
In each of the other scenarios at least one pairwise assessment of the null hypothesis was rejected.
As a whole, this data show that there are indeed variations in performance before considering the relative effect of virtualization.

Next we consider the pairwise evaluation of data from different scenarios (Tab. \ref{comparsions}).
In the experimental control samples (where there is no L3 virtualization), the null hypothesis was not rejected in any of the 16 comparisons.
When comparing the errors observed when running on the experiments on VirtualBox to those observed when running on Docker however, there were statistically significant differences.
On OSX, in more than half of the comparisons the null hypothesis was rejected.
When running on Ubuntu, the null hypothesis was only rejected in one of the 16 comparisons.
This suggests that experiments run in the two virtualization environments on Ubuntu are more consistent with each other than experiments run on OSX.

Finally, we compare the consistency of the four experiments using L3 virtualization with each of the two experiments where the technology was not used (the control scenarios).
As presented in Tab. \ref{comparsions2}, in both cases OSX is less consistent with the experimental control than Ubuntu.
Also, in both cases, Docker is less consistent with the control than experiments performed with VirtualBox.

The purpose of this work was to explore if there are differences that would be observed in the simulations executed through the use of different virtualization strategies, and the data shows this.
Moreover, the data show that the differences are conditioned on how virtualization is provided.
What is notable about these trends is that they are consistent over different infrastructures since there are different implementations of the virtualization tools for OSX and for Ubuntu.

\addtolength{\textheight}{-1cm}   

\section{CONCLUSIONS}
This work shows that there are quantifyable differences in results for robot simulations when virtualization is used to deploy them.
This means that conclusions drawn from such simulations are dependent on how and where they were deployed.
Our findings were acquired by implementing simulated robot control over a mininet-emulated network, and then deploying the simulation in VirtualBox, Docker and ``natively'' in three different operating environments.
The data we present show that the performance of VirtualBox is more consistent with non-virtualized environments than Docker.
Further, the data show that the variability in performance is dependent on the operating environment itself; e.g. comparisons on OSX are generally less consistent than Ubuntu.
At present, proponents of repeatable robotics simulation must be cautiously optimistic about both types of virtualization, and this work highlights the specific areas of concern.

The findings in this work are based on experiments requiring low bandwidth communication between controller and robot.
While the dynamics of the plant still resulted in a meaningful demonstration of control, future work will use higher bandwidth devices like cameras and depth sensors.
We will also extend analysis to include the effects of virtualization on the network packets that are used to transmit the control/sensor data.
Such extensions will provide the relevant insight, and data, to develop models that will better frame simulation results generated with virtualization.
More broadly, such models will also be key in compensating for the operation of simulation testbeds in heterogeneous computing environments.










\bibliographystyle{IEEEtran}

\end{document}